\pdfoutput=1

\documentclass[11pt]{article}

\usepackage[]{emnlp2022}

\usepackage{times}
\usepackage{latexsym}
\usepackage{bm}
\usepackage{comment}
\usepackage{upgreek}
\usepackage{tabularx}

\usepackage[T1]{fontenc}

\usepackage[utf8]{inputenc}

\usepackage{microtype}
\usepackage{enumerate}
\usepackage{enumitem}
\usepackage{color}
\usepackage{adjustbox}
\usepackage{booktabs}
\usepackage{makecell}
\usepackage{amsfonts}
\usepackage{amsthm}
\usepackage{amsmath}
\usepackage{amssymb}
\usepackage{multirow}
\usepackage{xfrac}
\usepackage{scalerel}
\usepackage{float}
\usepackage{inconsolata} 
\usepackage[belowskip=-10pt,aboveskip=10pt]{caption}
\usepackage{cleveref}

\usepackage{todonotes}
\makeatletter
\newcommand*\iftodonotes{\if@todonotes@disabled\expandafter\@secondoftwo\else\expandafter\@firstoftwo\fi}  
\makeatother

\renewcommand{\hat}{\widehat}

\newcommand{\WW}{\mathbf{W}}
\newcommand{\ww}{\mathbf{w}}
\newcommand{\ee}{\mathbf{e}}
\newcommand{\rr}{\mathbf{r}}
\newcommand{\tth}{\bm{\uptheta}}
\usepackage{mathtools}
\definecolor{darkgrey}{rgb}{0.2,0.2,0.2}
\definecolor{darkgreen}{rgb}{0.0,0.6,0.0}
\definecolor{darkblue}{rgb}{0.0,0.0,0.5}
\newcommand{\db}[1]{\textcolor{darkblue}{\bf\scriptstyle \selectfont \,(\pm#1)}}

\title{Eye-tracking based classification of Mandarin Chinese readers with and without dyslexia using neural sequence models}

\usepackage{tipa}

\newcommand{\uzh}{1}
\newcommand{\up}{2}
\newcommand{\ehk}{3}
\newcommand{\um}{4}

\author{Patrick Haller$^{\uzh}$,~\;~Andreas Säuberli$^{\uzh}$,~\;~Sarah E. Kiener$^{\uzh}$\\\textbf{Jinger Pan$^{\ehk}$,~\;~Ming Yan$^{\um}$,~\;~ Lena A. Jäger$^{\uzh,\up}$}\\
  $^{\uzh}$University of Zurich~\;~$^{\up}$University of Potsdam~\;~\\$^{\ehk}$The Education University of Hong Kong~\;~$^{\um}$University of Macau \\
 \texttt{\href{mailto:haller@cl.uzh.ch}{haller@cl.uzh.ch}}~\;~
   \texttt{\href{mailto:andreas@cl.uzh.ch}{andreas@cl.uzh.ch}}~\;~
     \texttt{\href{mailto:sarahelisabeth.kiener@uzh.ch}{sarahelisabeth.kiener@uzh.ch}}~\;~\\
       \texttt{\href{mailto:jpan@eduhk.hk}{jpan@eduhk.hk}}~\;~
         \texttt{\href{mailto:mingyan@um.edu.mo}{mingyan@um.edu.mo}}~\;~
  \texttt{\href{mailto:jaeger@cl.uzh.ch}{jaeger@cl.uzh.ch}}
}

\begin{document}

\maketitle
\begin{abstract}
Eye movements are known to reflect cognitive processes in reading, and psychological reading research has shown that eye gaze patterns differ between readers with and without dyslexia. In recent years, researchers have attempted to classify readers with dyslexia based on their eye movements using Support Vector Machines (SVMs). However, these approaches (i) are based on highly aggregated features averaged over all words read by a participant, thus disregarding the sequential nature of the eye movements, and (ii) do not consider the linguistic stimulus and its interaction with the reader’s eye movements. In the present work, we propose two simple sequence models that process eye movements on the entire stimulus without the need of aggregating features across the sentence. Additionally, we incorporate the linguistic stimulus into the model in two ways---contextualized word embeddings and manually extracted linguistic features. The models are evaluated on a Mandarin Chinese dataset containing eye movements from children with and without dyslexia. Our results show that (i) even for a logographic script such as Chinese, sequence models are able to classify dyslexia on eye gaze sequences, reaching state-of-the-art performance, and (ii) incorporating the linguistic stimulus does not help to improve classification performance.\footnote{Model code is publicly available and can be found under \url{https://github.com/hallerp/dyslexia-seqmod}.}
\end{abstract}

\section{Introduction}
\begin{figure}[h!]
    \centering
    \includegraphics[width=\linewidth]{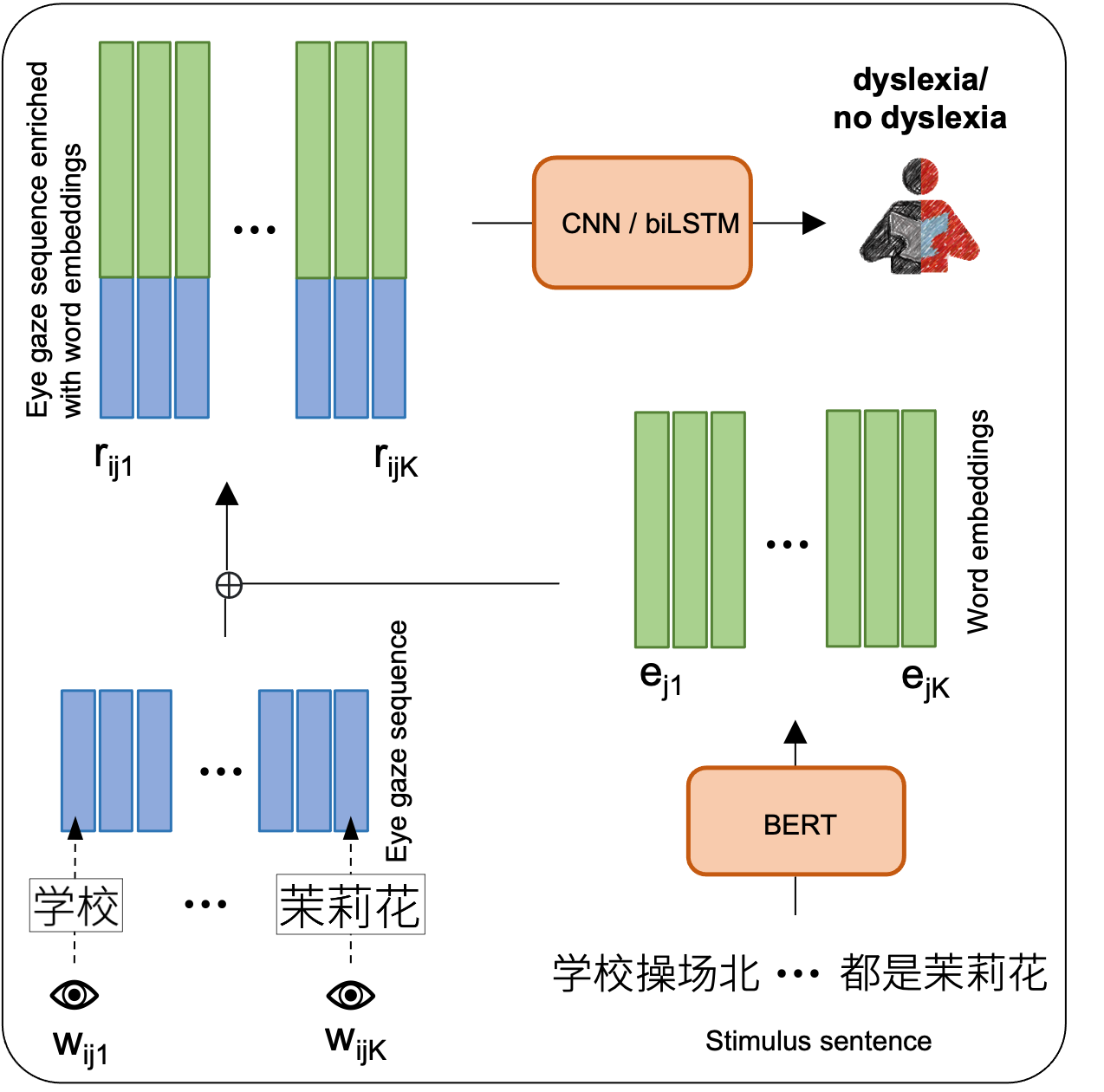}
    \caption{Proposed approach. Each eye-movement reading measure vector is concatenated with contextualized word embeddings and used as input for the sequence models to infer whether a reader suffers from dyslexia. \looseness=-1
    } \label{fig:model}
    \setlength{\belowcaptionskip}{-30pt}
\end{figure}
\label{sec:intro}
Reading effortlessly constitutes a key skill in modern society. Individuals suffering from developmental dyslexia are characterized by specific and persistent reading problems. Global prevalence estimates range from 3 to 7\% \citep{Landerl2013PredictorsComplexity, Peterson2012DevelopmentalDyslexia}. Previous research has consistently shown that early diagnosis and intervention is key to mitigate the resulting long-term consequences \citep{vaughn2010response}. \\
Psychological and clinical research on eye movement patterns has revealed that individuals with dyslexia exhibit gaze patterns that differ significantly from the patterns observed in  individuals without dyslexia \citep{Rayner1998EyeResearch, Pan2014Saccade-targetChinese}. In particular, scanpaths of individuals with dyslexia are characterized by longer fixation durations, more fixations, decreased saccade durations and a higher proportion of regressions. In recent years, increasing effort has been spent on utilizing these findings and applying supervised classification methods such as SVMs and Random Forests on eye movement data (see \citealt{Kaisar2020DevelopmentalSurvey} for an overview) to infer the presence or absence of dyslexia. There are several reasons why automatized approaches for assistance in dyslexia detection are desirable. Currently, paper-pencil diagnostic tools are conducted by trained speech therapists. These tools are time-intensive and are typically only considered after a suspected case has been reported by observant educational staff, leaving many cases overlooked. Eye-movement-based diagnostic tools have the potential to be deployed in schools in a relatively inexpensive manner and as part of a standard procedure aimed at early and comprehensive detection of dyslexia; making an important contribution to educational equity.\\
Although the aforementioned approaches provide promising results, they suffer from specific drawbacks: (i) The model input consists of eye movement features, aggregated for each subject over the presented stimulus material (text), thus disregarding the sequential nature of the eye movements; (ii) both the linguistic stimulus and its interaction with the reader’s eye movements are not considered. For classification purposes, this does not pose a problem \emph{per se}. However, it does not allow us to investigate questions such as: Which words (or, more specifically, what linguistic properties of the stimulus) are particularly informative to discriminate between individuals with and without dyslexia?\\
In the present work, we propose two neural sequence models, depicted in Figure~\ref{fig:model}, that process the eye movements on the entire stimulus without the necessity of feature aggregation over the sentence. To incorporate the linguistic stimulus into the model, we use pre-trained contextualized word embeddings. We evaluate our model on an eye-tracking-while-reading dataset from children with and without dyslexia reading Mandarin Chinese sentences by \citet{Pan2014Saccade-targetChinese}.

\section{Related Work}
\subsection{ML-based detection of dyslexia}
\label{sec:related-work}
To date, various data types and signals have been utilized to solve the task of automated detection of dyslexia such as text, MRI scans \citep{cui2016disrupted}, EEG recordings \citep{frid2012svm}, student engagement data \citep{abdul2018dyslexia} as well as eye-tracking data \citep{Rello2015DetectingMeasures, Raatikainen2021DetectionData, Benfatto2016ScreeningReading}.
\citet{Benfatto2016ScreeningReading} train a \textit{Support Vector Machine with recursive feature elimination} (SVM-RFE) on 168 eye-tracking features obtained from an eye-tracking-while-reading dataset from 185 Swedish children (aged 9-10 years). Their best SVM-RFE model selected 48 features and achieved an accuracy score of 95.6\% $\pm$ 4.5\% (sic!) on a balanced dataset. We reimplement this method and use it as a reference method (cf.~\ref{baseline}).
\citet{JothiPrabha2020PredictiveEvents}, using the same dataset as \citet{Benfatto2016ScreeningReading}, experiment with various feature selection algorithms and machine learning models. They find that feature selection via Principle Component Analysis (PCA) in combination with a Particle Swarm Optimization based Hybrid Kernel SVM classifier yields the best accuracy.\\
\citet{Raatikainen2021DetectionData} combine a Random Forest classifier for feature selection with an SVM, achieving an accuracy of 89.7\%. They expand their feature space with transition matrices that represent the number of transitions between the different segments (question, answer selection) in a trial as well as the number of gaze shifts within one segment. 

\subsection{Modeling eye-tracking data with deep neural sequence models}
\paragraph{Eye movement data for task inference.} Deep neural sequence models have been deployed to solve inference tasks based on eye movements such as reader~\citep{jaeger2020deep} and viewer identification \citep{Lohr2020ABiometrics,Makowski2020BiometricEyes, Makowski2021DeepEyedentificationLive:Networks}, ADHD detection \citep{deng2022detection} as well as the prediction of reading comprehension~\citep{reich2022inferring}. 
\paragraph{Integrating the linguistic stimulus.} There has been growing interest in combining language and eye movement models to predict gaze patterns during naturalistic reading \citep{hollenstein2021reading, merkx-frank-2021-human,hollenstein2022patterns}. \citet{wiechmann-kerz-2022-measuring} investigate the role of general text features and their interaction with eye movement patterns in predicting human reading behavior and find that models incorporating the linguistic stimulus improves prediction accuracy.

\section{Problem Setting}
We investigate the two closely related tasks of classifying (i) whether a given eye gaze sequence on one sentence is from a reader with or without dyslexia and (ii) whether a given eye gaze sequence on a set of sentences is from a reader with or without dyslexia. Formally, our training data can be represented as a set $\mathcal{D} = \{(\WW_{11}, y_{1}), \dots, (\WW_{NM}, y_{N})\}$, where $\WW_{ij}=\langle \ww_{ij1} \dots \ww_{ijK} \rangle$ is a sequence of reading measure vectors\footnote{Cf. the list of reading measures in Appendix~\ref{sec:appendix:rm}.} for each word $k \in 1\dots K_j$ obtained from subject $i$ reading sentence $j$, where $N$ is the number of participants, $M$ is the number of stimulus sentences read by each of the participants and $K_j$ the number of words in a given sentence $j$. Each reading measure vector consists of $R$ reading measures, i.e., $\ww_{ijk}= (r_{ijk1} \dots r_{ijkR})$. The binary target label $y_i$ denotes whether participant $i$ is a reader with or without dyslexia. For (i), our goal is to train a binary classifier $g_{\tth}$ such that 
$$\hat{y}_{i}=\begin{cases}
1, \quad \text{if } g_{\tth}(\WW_{ij})\geq \delta \\
0, \quad \text{else},
\end{cases}$$ where $\delta$ denotes the decision threshold and $\tth$ the set of hyperparameters. Accordingly, for (ii), $\hat{y}_i=1, \text{if } \frac{1}{M} \sum_{j=1}^{M} g_{\tth}(\WW_{ij})\geq \delta$. \\
 The performance of a binary model can be characterized by a false-positive and a true-positive rate. By altering the decision threshold $\delta$, a receiver operator characteristic (ROC) curve can be derived, with the area under the curve providing an aggregated measure for all possible values of $\delta$.

\section{Methods}
\subsection{Reference method} \label{baseline}
As a baseline method, we train an SVM-RFE, following the procedure described by \citet{Benfatto2016ScreeningReading}. We use the \emph{scikit-learn} implementation \citep{scikit-learn} of the SVM-RFE with a linear kernel. In the \textit{subject-prediction} setting, we use eye movement features from each subject aggregated (mean and standard deviation)  across trials and sentences as input vectors. In the \textit{sentence-prediction} setting, we use aggregates of each sentence over all trials, yielding $2\times12=24$ features per instance in both settings.\footnote{We also experimented with training random forests as baseline, however, they were outperformed by the SVM-RFE.}

\subsection{Proposed neural sequence models}
Both models take as input an enriched reading measure vector $\rr_{ij}$ (cf. Section \ref{sec:we}) of a sentence $j$ read by participant $i$, normalized for each train/test set separately, and predict a label $y_i$. We tune both models using random search.
\paragraph{LSTM.}
We implement a bidirectional recurrent neural network with LSTM cells. The mean of the hidden states is fed into a linear layer projecting it down to a single sigmoid output to represent the label prediction. Optimized hyperparameters and search space are reported in Appendix \ref{tab:hyperparameter-space}.

\paragraph{CNN.} We implement a CNN that convolves the input accross the word sequence axis. It consists of two convolutional layers, each followed by a pooling layer, two dense layers, and a sigmoid output unit. Hyperparameters are listed in Appendix~\ref{tab:hyperparameter-space}.

\subsubsection{Incorporating the linguistic stimulus}
\paragraph{Using contextualized word embeddings.}\label{sec:we}
To incorporate the linguistic stimulus (the words occurring in the current sentence), we first extract 768-dimensional BERT embeddings $\ee_{jk}$ for each word $w$ in a given sentence $j$, using the pre-trained BERT\textsubscript{\textsc{base}}-embeddings, provided by Hugging Face \citep{wolf2020transformers}, and concatenate them with the reading measure vector $\ww_{ijk}$, resulting in an enriched reading measure vector $\rr_{ijk}$. \label{sec:mde} Concatenating the full embedding to the feature vectors results in $768+R$ dimensions, resulting in a substantial increase in parameters to be estimated. Given the small amount of available training data, we test two methods of dimensionality reduction: (i) We perform PCA on the word embeddings and use the first 20 principal components. (ii) \emph{Mean-difference-encoding}: In order to capture domain-specific information from the word embeddings relating to differences in reading behaviour exhibited by individuals with and without dyslexia, we propose an alternative method, which we call \emph{mean-difference-encoding}: We train a feed-forward neural network with one hidden layer of size 20 to predict differences between the mean values of each eye movement feature between the two groups for each word based on its original word embedding. The values of the hidden layer are a compressed representation of the original embedding that is optimized to encode information that discriminates between children with and without dyslexia. In order to avoid train-test data leakage, in each fold, the mean-difference-encoder is trained from scratch on the respective training set.
\paragraph{Using manually extracted features.}
As an alternative way to incorporate the linguistic stimulus, we add a range of \emph{manually extracted linguistic features} for each token $w_{jk}$ in sentence $j$: Surprisal, i.e., $- \log p(w_{jk} \mid \mathbf w_{j< k})$, estimated with GPT-2~\citep{radford2019language}, part of speech, dependency relation type, distance to syntactic head, extracted using spaCy~\citep{honnibal2020spacy}, mean character frequency and lexical frequency extracted from SUBTLEX-CH \citep{cai2010subtlex}.

\section{Experiments}
\paragraph{Data.}
We employ eye-tracking-while-reading data from 62 Mandarin Chinese children (33 with dyslexia) provided by \citet{Pan2014Saccade-targetChinese}. Participants were instructed to read 60 sentences out loud while their eye movements were recorded. 40 sentences were selected from fifth grade textbooks and 20 additional control sentences were extracted from the Beijing Sentence Corpus \citep{Pan2021TheNorms}. The dyslexia label had been assigned when a child scored at least 1.5 standard deviations below their corresponding age mean in standard character recognition test \citep{shu2003properties}.

\subsection{Evaluation procedure} \label{sec:evaluation}
We evaluate our models using 10-fold nested cross-validation in two settings. In the \textit{sentence prediction} setting, we predict the label from a single sentence, read by a given subject. In the \textit{subject prediction} setting, we average the sigmoid outputs from all sentences read by a given subject in order to obtain a subject-level prediction. In both settings, sentences are stratified over 10 folds, balanced by group. Data from the same subject is always constrained to one fold, thus, the model always makes predictions for unseen subjects.
\paragraph{Hyperparameter tuning.} For each test fold, we iterate through 9 validation folds, training 50 LSTM and 100 CNN models using randomly sampled parameter combinations for each fold. We select the highest scoring parameter set over all 9 validation folds and train a final model using 8 training folds. We use one left-out fold for early stopping and evaluate it on the test fold.

\begin{table*}
  \centering
  \small
  \begin{adjustbox}{max width=\textwidth}
  \begin{tabular}{lll|rrrrr}
   \toprule
   \multicolumn{3}{c}{Architecture} & \multicolumn{5}{c}{Evaluation Metrics} \\
 & {Model} & {Stimulus representation} & \textsc{auc} & Accuracy & Recall & Precision & $F_1$ \\
     \hline\\[-1.5ex]
     \parbox[t]{2mm}{\multirow{9}{*}{\rotatebox[origin=c]{90}{\textsc{subject-level}}}}
& \emph{Baseline} &  & $\mathbf{0.93\db{0.03}}$ & $\mathbf{0.90\db{0.03}}$ & $0.87\db{0.04}$ & $0.97\db{0.03}$ & $0.91\db{0.03}$\vspace{1mm} \\ 
     & \multirow{4}{*}{\emph{LSTM}}
    & None & $0.91\db{0.03}$ & $\mathbf{0.90\db{0.03}}$ & $\mathbf{0.88\db{0.05}}$ & $0.98\db{0.02}$ & $\mathbf{0.92\db{0.03}}$ \\
    && BERT meandiff & $0.88\db{0.03}$ & $0.80\db{0.06}$ & $0.78\db{0.06}$ & $0.93\db{0.06}$ & $0.83\db{0.06}$ \\
    && BERT PCA & $0.90\db{0.03}$ & $0.83\db{0.05}$ & $0.81\db{0.06}$ & $0.97\db{0.03}$ & $0.87\db{0.04}$ \\
    && Manually extracted & $0.87\db{0.04}$ & $0.87\db{0.05}$ & $0.84\db{0.05}$ & $0.97\db{0.03}$ & $0.89\db{0.04}$\vspace{1mm} \\
   & \multirow{4}{*}{\emph{CNN}}
    & None & $0.91\db{0.04}$ & $0.90\db{0.03}$ & $0.86\db{0.04}$ & $\mathbf{1.00\db{0.00}}$ & $\mathbf{0.92\db{0.02}}$ \\
    && BERT meandiff & $0.91\db{0.03}$ & $0.90\db{0.03}$ & $0.88\db{0.04}$ & $0.97\db{0.03}$ & $0.91\db{0.02}$ \\
    && BERT PCA & $\mathbf{0.93\db{0.03}}$ & $0.87\db{0.02}$ & $0.86\db{0.04}$ & $0.93\db{0.04}$ & $0.88\db{0.02}$ \\
    && Manually extracted & $0.89\db{0.04}$ & $0.83\db{0.04}$ & $0.80\db{0.05}$ & $0.97\db{0.03}$ & $0.86\db{0.03}$ \\
    \midrule
         \parbox[t]{2mm}{\multirow{9}{*}{\rotatebox[origin=c]{90}{\textsc{sentence-level}}}}
&    \emph{Baseline} &  & $\mathbf{0.85\db{0.03}}$ & $\mathbf{0.78\db{0.02}}$ & $\mathbf{0.79\db{0.04}}$ & $0.76\db{0.02}$ & $0.77\db{0.02}$\vspace{1mm} \\ 
  & \multirow{4}{*}{\emph{LSTM}}
     & None & $\mathbf{0.85\db{0.03}}$ & $0.77\db{0.03}$ & $0.74\db{0.04}$ & $0.83\db{0.03}$ & $\mathbf{0.78\db{0.03}}$ \\
     && BERT meandiff & $0.81\db{0.04}$ & $0.68\db{0.04}$ & $0.65\db{0.04}$ & $\mathbf{0.86\db{0.05}}$ & $0.72\db{0.03}$ \\
     && BERT PCA  & $0.79\db{0.04}$ & $0.66\db{0.04}$ & $0.64\db{0.04}$ & $0.85\db{0.05}$ & $0.71\db{0.03}$ \\
     && Manually extracted & $0.77\db{0.05}$ & $0.71\db{0.03}$ & $0.67\db{0.03}$ & $0.85\db{0.05}$ & $0.74\db{0.03}$\vspace{1mm} \\
   & \multirow{4}{*}{\emph{CNN}}
    & None & $0.84\db{0.02}$ & $0.76\db{0.02}$ & $0.73\db{0.02}$ & $0.83\db{0.04}$ & $0.77\db{0.02}$ \\
    && BERT meandiff & $0.82\db{0.03}$ & $0.75\db{0.02}$ & $0.72\db{0.02}$ & $0.82\db{0.04}$ & $0.76\db{0.02}$ \\
    && BERT PCA & $0.82\db{0.03}$ & $0.74\db{0.02}$ & $0.70\db{0.02}$ & $0.85\db{0.04}$ & $0.76\db{0.02}$ \\
    && Manually extracted & $0.82\db{0.03}$ & $0.74\db{0.02}$ & $0.69\db{0.02}$ & $\mathbf{0.86\db{0.03}}$ & $0.76\db{0.02}$ \\
    \bottomrule
  \end{tabular} 
  \end{adjustbox}
  \caption{Classification results using 10-fold cross validation on subject- and sentence-level. We report \textsc{auc}, accuracy, recall, precision and $F1$ [results ± standard error]. The latter four were computed for a decision threshold of $0.5$. \looseness=-1}
  \label{tab:results-subj}
\end{table*}

\subsection{Results}

\begin{figure}
    \centering
    \includegraphics[width=\linewidth]{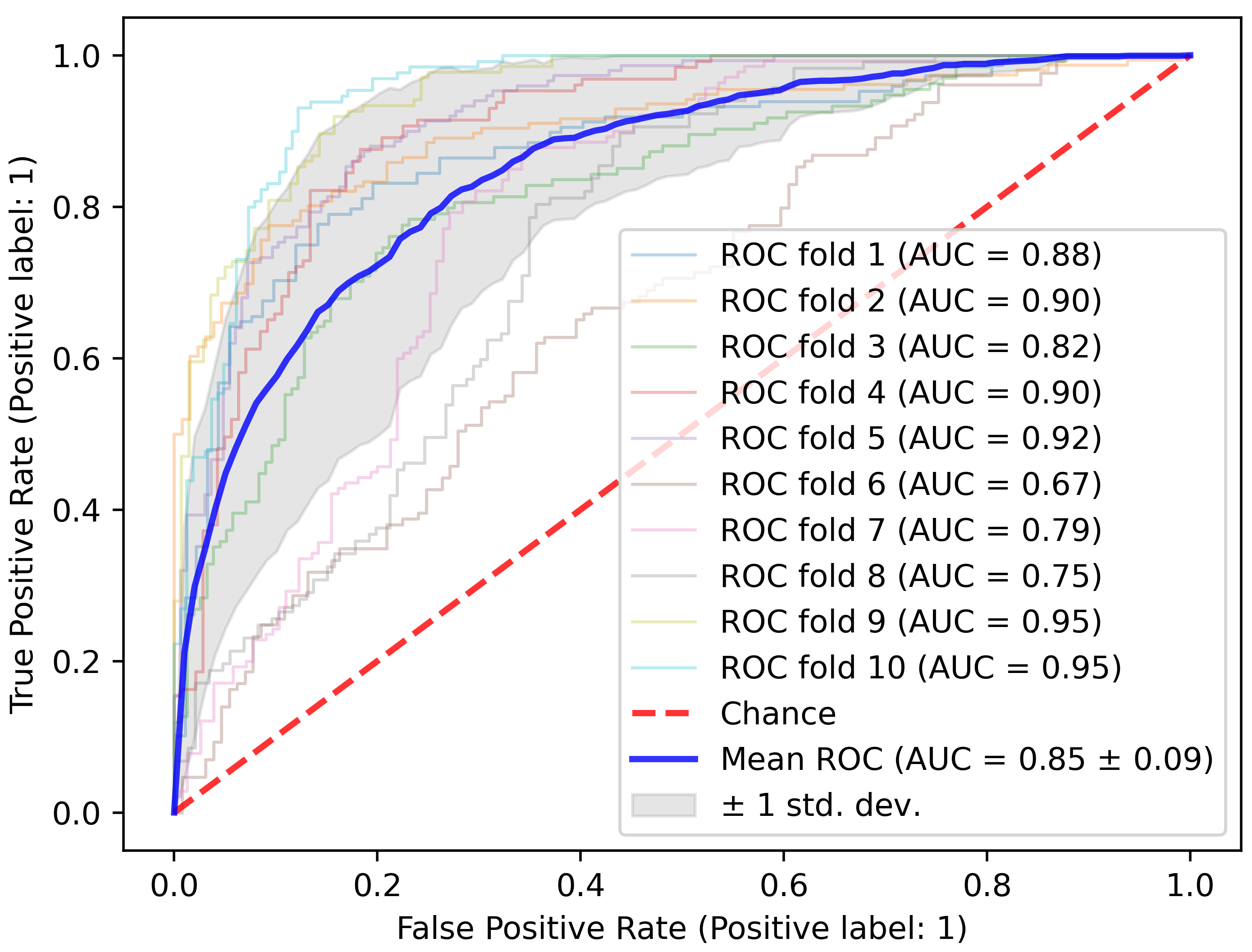}
    \caption{ROC curves over all test sets for best performing model (LSTM with no linguistic stimulus representation) on sentence-level. \looseness=-1
    } \label{fig:roc}
    \setlength{\belowcaptionskip}{-30pt}
\end{figure}

For all methods, we report \textsc{auc} as well as accuracy, recall, precision and the harmonic precision-recall mean $F_1$ for a decision threshold of $0.5$ on subject- and sentence-level. As can be seen in Table \ref{tab:results-subj}, our proposed models reach but do not outperform state-of-the art performance. While on subject-level, the CNN architecture enriched with PCA-reduced word embeddings achieves the highest \textsc{auc}, on sentence-level, the best results are obtained by the LSTM that solely includes eye-movement features. Overall, we note that classification performance on subject-level is higher than on sentence-level and that adding the linguistic stimulus does not aid classification performance, neither as contextualized word embeddings nor as manually-extracted features. Furthermore, as can be seen in Figure~\ref{fig:roc}, performance varies considerably with respect to different test sets. We also observe that the variance in AUC for models enriched with the linguistic stimulus is larger for LSTMs compared to CNNs. Lastly, our domain-specific dimensionality reduction method (cf. Section~\ref{sec:mde}) has no advantage over PCA, although the former is explicitly trained on differences between the two groups.

\section{Discussion}
Our proposed neural sequence models reach state-of-the-art performance on solving the task of detecting dyslexia from eye gaze sequences, for the first time investigated for a logographic script such as Chinese. Our results suggest that for our dataset, (i) neural architectures processing eye-movement sequences along the sentence have no advantage over the parsimonious SVM-baseline where features are aggregated over the sentence, and (ii) enabling the interaction between stimulus input and eye movements does not improve classification performance. However, after having shown that our approach is able to reach SOTA performance, we aim to exploit its properties to investigate the informativeness of particular sentences, words, and other linguistic sub-units for dyslexia detection in the future.\\
Furthermore, for all investigated models, the overall performance appears to be driven by a small subset of individuals who presumably exhibit less typical reading behavior among their group and were more difficult to classify. Given that dyslexia is a spectrum disorder---not binary as it is often perceived---it is to be expected that individuals that are not located at the two extremes (clearly dyslexic or clearly not dyslexic) are more difficult to classify in a binary environment.\\
Our study was able to show that an SVM-based approach, previously applied to alphabetic languages such as Swedish and Spanish, also works well on a logographic script such as Chinese. In future work, we would like to test our approach on alphabetic language data sets. This is particularly interesting given the fact that young Chinese readers are faced with different challenges, e.g., the absence of orthographic word boundaries, therefore requiring word segmentation, and the much larger number of characters required to be memorized.

\paragraph{Limitations.} 
It should be noted that our dataset contained very little data. Considering that the number of parameters of our sequence models exceeded the one of the baseline model by orders of magnitude, it might be worth comparing the approaches again, once more data is available. The problem of data scarcity might be alleviated by pre-training on domain general eye-tracking datasets or with data augmentation methods\footnote{In a preliminary experiment, we pre-trained our models on the Beijing Sentence Corpus \citep{Pan2021TheNorms} and found that it did not increase classification performance.}. Furthermore, we did not have access to the raw scores of the character recognition task. While our methods did not outperform the baseline in this binary environment, it would be interesting to assess their performance on a regression task.

\section{Conclusion}
For the first time, we deploy models to detect dyslexia from eye gaze sequences on data from Mandarin Chinese readers. We propose two sequence classification approaches that (i) take as input the full, non-aggregated linguistic stimulus and (ii) model the interaction of the stimulus with the eye movements. As a comparison, we adapt a previously proposed SVM-based approach for Mandarin Chinese. We find that all models reach SOTA performance for data based on a logographic script such as Chinese. In addition, we find that incorporating the linguistic stimulus does not improve the models' performance. Given that we reach SOTA performance on a very small dataset, our approach has proven worthwhile to be pursued, expanded, and further tested (e.g., on alphabetic language data sets). It has the potential to be successfully deployed in the context of automatized approaches for dyslexia detection with the final objective being the improvement of educational equity.

\section*{Acknowledgments}
We thank David Reich for his continuous feedback on the model architectures as well as our anonymous reviewers for their invaluable feedback on the manuscript for this work. Lena J\"ager was partially funded by the German Federal Ministry of Education and Research under grant 01\textpipe S20043.

\bibliography{anthology,custom,acl_latex}

\begin{thebibliography}{30}
\expandafter\ifx\csname natexlab\endcsname\relax\def\natexlab#1{#1}\fi

\bibitem[{Abdul~Hamid et~al.(2018)Abdul~Hamid, Admodisastro, Manshor,
  Kamaruddin, and Ghani}]{abdul2018dyslexia}
Siti~Suhaila Abdul~Hamid, Novia Admodisastro, Noridayu Manshor, Azrina
  Kamaruddin, and Abdul Azim~Abd Ghani. 2018.
\newblock \href {https://doi.org/10.1007/978-3-319-72550-5_36} {Dyslexia
  adaptive learning model: Student engagement prediction using machine learning
  approach}.
\newblock In \emph{International Conference on Soft Computing and Data Mining},
  pages 372--384. Springer.

\bibitem[{Benfatto et~al.(2016)Benfatto, Seimyr, Ygge, Pansell, Rydberg, and
  Jacobson}]{Benfatto2016ScreeningReading}
Mattias~Nilsson Benfatto, Gustaf~Öqvist Seimyr, Jan Ygge, Tony Pansell, Agneta
  Rydberg, and Christer Jacobson. 2016.
\newblock \href {https://doi.org/10.1371/journal.pone.0165508} {{Screening for
  dyslexia using eye tracking during reading}}.
\newblock \emph{PLoS ONE}, 11(12):e0165508.

\bibitem[{Cai and Brysbaert(2010)}]{cai2010subtlex}
Qing Cai and Marc Brysbaert. 2010.
\newblock \href {https://doi.org/10.1371/journal.pone.0010729} {{SUBTLEX-CH:
  Chinese word and character frequencies based on film subtitles}}.
\newblock \emph{PLoS ONE}, 5(6):e10729.

\bibitem[{Cui et~al.(2016)Cui, Xia, Su, Shu, and Gong}]{cui2016disrupted}
Zaixu Cui, Zhichao Xia, Mengmeng Su, Hua Shu, and Gaolang Gong. 2016.
\newblock \href {https://doi.org/10.1002/hbm.23112} {Disrupted white matter
  connectivity underlying developmental dyslexia: A machine learning approach}.
\newblock \emph{Human Brain Mapping}, 37(4):1443--1458.

\bibitem[{Deng et~al.(2022)Deng, Prasse, Reich, Dziemian,
  Stegenwallner-Sch{\"u}tz, Krakowczyk, Makowski, Langer, Scheffer, and
  J{\"a}ger}]{deng2022detection}
Shuwen Deng, Paul Prasse, David~R. Reich, Sabine Dziemian, Maja
  Stegenwallner-Sch{\"u}tz, Daniel Krakowczyk, Silvia Makowski, Nicolas Langer,
  Tobias Scheffer, and Lena~A. J{\"a}ger. 2022.
\newblock \href {https://doi.org/10.48550/arXiv.2207.01377} {Detection of
  {ADHD} based on eye movements during natural viewing}.
\newblock In \emph{Joint European Conference on Machine Learning and Knowledge
  Discovery in Databases}.

\bibitem[{Frid and Breznitz(2012)}]{frid2012svm}
Alex Frid and Zvia Breznitz. 2012.
\newblock \href {https://doi.org/10.1109/EEEI.2012.6377068} {{An SVM based
  algorithm for analysis and discrimination of dyslexic readers from regular
  readers using ERPs}}.
\newblock In \emph{2012 IEEE 27th Convention of Electrical and Electronics
  Engineers in Israel}, pages 1--4. Institute of Electrical and Electronics
  Engineers.

\bibitem[{Hollenstein et~al.(2022)Hollenstein, Gonzalez-Dios, Beinborn, and
  Jäger}]{hollenstein2022patterns}
Nora Hollenstein, Itziar Gonzalez-Dios, Lisa Beinborn, and Lena~A. Jäger.
  2022.
\newblock Patterns of text readability in human and predicted eye movements.
\newblock In \emph{Proceedings of the Workshop on the Cognitive Aspects of the
  Lexicon}. Association for Computational Linguistics.

\bibitem[{Hollenstein et~al.(2021)Hollenstein, Tr{\"{o}}ndle, Plomecka,
  Kiegeland, {\"{O}}zyurt, J{\"{a}}ger, and Langer}]{hollenstein2021reading}
Nora Hollenstein, Marius Tr{\"{o}}ndle, Martyna Plomecka, Samuel Kiegeland,
  Yilmazcan {\"{O}}zyurt, Lena~A. J{\"{a}}ger, and Nicolas Langer. 2021.
\newblock \href {https://doi.org/10.48550/arXiv.2112.06310} {{Reading task
  classification using EEG and eye-tracking data}}.
\newblock \emph{arXiv:2112.06310}.

\bibitem[{Honnibal et~al.(2020)Honnibal, Montani, Van~Landeghem, and
  Boyd}]{honnibal2020spacy}
Matthew Honnibal, Ines Montani, Sofie Van~Landeghem, and Adriane Boyd. 2020.
\newblock \href {https://spacy.io/} {{spaCy}: Industrial-strength natural
  language processing in {P}ython}.
\newblock Zenodo.
\newblock {https://spacy.io/}.

\bibitem[{J{\"{a}}ger et~al.(2019)J{\"{a}}ger, Makowski, Prasse, Liehr,
  Seidler, and Scheffer}]{jaeger2020deep}
Lena~A. J{\"{a}}ger, Silvia Makowski, Paul Prasse, Sascha Liehr, Maximilian
  Seidler, and Tobias Scheffer. 2019.
\newblock \href {https://doi.org/10.1007/978-3-030-46147-8_18} {{Deep
  Eyedentification: Biometric identification using micro-movements of the
  eye}}.
\newblock In \emph{Joint European Conference on Machine Learning and Knowledge
  Discovery in Databases}, Lecture Notes in Computer Science, pages 299--314.
  Springer.

\bibitem[{Jothi~Prabha and Bhargavi(2020)}]{JothiPrabha2020PredictiveEvents}
Appadurai Jothi~Prabha and Renta Bhargavi. 2020.
\newblock \href {https://doi.org/10.1016/j.cmpb.2020.105538} {Predictive model
  for dyslexia from fixations and saccadic eye movement events}.
\newblock \emph{Computer Methods and Programs in Biomedicine}, 195:105538.

\bibitem[{Kaisar(2020)}]{Kaisar2020DevelopmentalSurvey}
Shahriar Kaisar. 2020.
\newblock \href {https://doi.org/10.1016/j.icte.2020.05.006} {{Developmental
  dyslexia detection using machine learning techniques: A survey}}.
\newblock \emph{ICT Express}, 6(3):181--184.

\bibitem[{Landerl et~al.(2013)Landerl, Ramus, Moll, Lyytinen, Lepp{\"{a}}nen,
  Lohvansuu, O'Donovan, Williams, Bartling, Bruder, Kunze, Neuhoff, Tõth,
  Honbolygõ, Cs{\'{e}}pe, Bogliotti, Iannuzzi, Chaix, D{\'{e}}monet, Longeras,
  Valdois, Chabernaud, Delteil-Pinton, Billard, George, Ziegler, Comte-Gervais,
  Soares-Boucaud, G{\'{e}}rard, Blomert, Vaessen, Gerretsen, Ekkebus, Brandeis,
  Maurer, Schulz, Van Der~Mark, M{\"{u}}ller-Myhsok, and
  Schulte-K{\"{o}}rne}]{Landerl2013PredictorsComplexity}
Karin Landerl, Franck Ramus, Kristina Moll, Heikki Lyytinen, Paavo~H.T.
  Lepp{\"{a}}nen, Kaisa Lohvansuu, Michael O'Donovan, Julie Williams, Jürgen
  Bartling, Jennifer Bruder, Sarah Kunze, Nina Neuhoff, Dénes Tõth, Ferenc
  Honbolygõ, Valéria Cs{\'{e}}pe, Caroline Bogliotti, Stéphanie Iannuzzi,
  Yves Chaix, Jean~François D{\'{e}}monet, Emilie Longeras, Sylviane Valdois,
  Camille Chabernaud, Florence Delteil-Pinton, Catherine Billard, Florence
  George, Johannes~C. Ziegler, Isabelle Comte-Gervais, Isabelle Soares-Boucaud,
  Christophe~Loïc G{\'{e}}rard, Leo Blomert, Anniek Vaessen, Patty Gerretsen,
  Michel Ekkebus, Daniel Brandeis, Urs Maurer, Enrico Schulz, Sanne Van
  Der~Mark, Bertram M{\"{u}}ller-Myhsok, and Gerd Schulte-K{\"{o}}rne. 2013.
\newblock \href {https://doi.org/10.1111/jcpp.12029} {{Predictors of
  developmental dyslexia in European orthographies with varying complexity}}.
\newblock \emph{Journal of Child Psychology and Psychiatry and Allied
  Disciplines}, 54(6):686--694.

\bibitem[{Lohr et~al.(2020)Lohr, Griffith, Aziz, and
  Komogortsev}]{Lohr2020ABiometrics}
Dillon Lohr, Henry Griffith, Samantha Aziz, and Oleg Komogortsev. 2020.
\newblock \href {https://doi.org/10.1109/IJCB48548.2020.9304859} {A metric
  learning approach to eye movement biometrics}.
\newblock In \emph{2020 IEEE International Joint Conference on Biometrics},
  pages 1--7. Institute of Electrical and Electronics Engineers.

\bibitem[{Makowski et~al.(2020)Makowski, J{\"a}ger, Prasse, and
  Scheffer}]{Makowski2020BiometricEyes}
Silvia Makowski, Lena~A. J{\"a}ger, Paul Prasse, and Tobias Scheffer. 2020.
\newblock \href {https://doi.org/10.1109/IJCB48548.2020.9304900} {Biometric
  identification and presentation-attack detection using micro- and
  macro-movements of the eyes}.
\newblock In \emph{2020 IEEE International Joint Conference on Biometrics},
  pages 1--10. Institute of Electrical and Electronics Engineers.

\bibitem[{Makowski et~al.(2021)Makowski, Prasse, Reich, Krakowczyk, J{\"a}ger,
  and Scheffer}]{Makowski2021DeepEyedentificationLive:Networks}
Silvia Makowski, Paul Prasse, David~R. Reich, Daniel Krakowczyk, Lena~A.
  J{\"a}ger, and Tobias Scheffer. 2021.
\newblock \href {https://doi.org/10.1109/TBIOM.2021.3116875}
  {{DeepEyedentificationLive: Oculomotoric biometric identification and
  presentation-attack detection using deep neural networks}}.
\newblock \emph{IEEE Transactions on Biometrics, Behavior, and Identity
  Science}, 3(4):506--518.

\bibitem[{Merkx and Frank(2021)}]{merkx-frank-2021-human}
Danny Merkx and Stefan~L. Frank. 2021.
\newblock \href {https://doi.org/10.18653/v1/2021.cmcl-1.2} {Human sentence
  processing: Recurrence or attention?}
\newblock In \emph{Proceedings of the Workshop on Cognitive Modeling and
  Computational Linguistics}, pages 12--22. Association for Computational
  Linguistics.

\bibitem[{Pan et~al.(2014)Pan, Yan, Laubrock, Shu, and
  Kliegl}]{Pan2014Saccade-targetChinese}
Jinger Pan, Ming Yan, Jochen Laubrock, Hua Shu, and Reinhold Kliegl. 2014.
\newblock \href {https://doi.org/10.1016/j.visres.2014.01.014} {{Saccade-target
  selection of dyslexic children when reading Chinese}}.
\newblock \emph{Vision Research}, 97:24--30.

\bibitem[{Pan et~al.(2022)Pan, Yan, Richter, Shu, and Kliegl}]{Pan2021TheNorms}
Jinger Pan, Ming Yan, Eike~M. Richter, Hua Shu, and Reinhold Kliegl. 2022.
\newblock \href {https://doi.org/10.3758/s13428-021-01730-2} {{The Beijing
  Sentence Corpus: A Chinese sentence corpus with eye movement data and
  predictability norms}}.
\newblock \emph{Behavior Research Methods}, 54(4):1989--2000.

\bibitem[{Pedregosa et~al.(2011)Pedregosa, Varoquaux, Gramfort, Michel,
  Thirion, Grisel, Blondel, Prettenhofer, Weiss, Dubourg, Vanderplas, Passos,
  Cournapeau, Brucher, Perrot, and Duchesnay}]{scikit-learn}
Fabian Pedregosa, Ga{\"{e}}l Varoquaux, Alexandre Gramfort, Vincent Michel,
  Bertrand Thirion, Olivier Grisel, Mathieu Blondel, Peter Prettenhofer, Ron
  Weiss, Vincent Dubourg, Jake Vanderplas, Alexandre Passos, David Cournapeau,
  Matthieu Brucher, Matthieu Perrot, and Edouard Duchesnay. 2011.
\newblock \href {https://doi.org/10.48550/arXiv.1201.0490} {Scikit-learn:
  Machine learning in {P}ython}.
\newblock \emph{Journal of Machine Learning Research}, 12:2825--2830.

\bibitem[{Peterson and Pennington(2012)}]{Peterson2012DevelopmentalDyslexia}
Robin~L. Peterson and Bruce~F. Pennington. 2012.
\newblock \href {https://doi.org/10.1016/S0140-6736(12)60198-6} {{Developmental
  dyslexia}}.
\newblock \emph{The Lancet}, 379(9830):1997--2007.

\bibitem[{Raatikainen et~al.(2021)Raatikainen, Hautala, Loberg,
  K{\"{a}}rkk{\"{a}}inen, Lepp{\"{a}}nen, and
  Nieminen}]{Raatikainen2021DetectionData}
Peter Raatikainen, Jarkko Hautala, Otto Loberg, Tommi K{\"{a}}rkk{\"{a}}inen,
  Paavo Lepp{\"{a}}nen, and Paavo Nieminen. 2021.
\newblock \href {https://doi.org/10.1016/j.array.2021.100087} {{Detection of
  developmental dyslexia with machine learning using eye movement data}}.
\newblock \emph{Array}, 12:100087.

\bibitem[{Radford et~al.(2019)Radford, Wu, Child, Luan, Amodei, and
  Sutskever}]{radford2019language}
Alec Radford, Jeff Wu, Rewon Child, David Luan, Dario Amodei, and Ilya
  Sutskever. 2019.
\newblock \href {https://openai.com/blog/better-language-models/} {Language
  models are unsupervised multitask learners}.
\newblock \emph{OpenAI Blog}, 1(8):9.

\bibitem[{Rayner(1998)}]{Rayner1998EyeResearch}
Keith Rayner. 1998.
\newblock \href {https://doi.org/10.1037/0033-2909.124.3.372} {Eye movements in
  reading and information processing: 20 years of research}.
\newblock \emph{Psychological Bulletin}, 124(3):372--422.

\bibitem[{Reich et~al.(2022)Reich, Prasse, Tschirner, Haller, Goldhammer, and
  J{\"a}ger}]{reich2022inferring}
David~R. Reich, Paul Prasse, Chiara Tschirner, Patrick Haller, Frank
  Goldhammer, and Lena~A. J{\"a}ger. 2022.
\newblock \href {https://doi.org/10.1145/3517031.3529639} {Inferring native and
  non-native human reading comprehension and subjective text difficulty from
  scanpaths in reading}.
\newblock In \emph{2022 Symposium on Eye Tracking Research and Applications},
  ETRA '22, pages 1--8. Association for Computing Machinery.

\bibitem[{Rello and Ballesteros(2015)}]{Rello2015DetectingMeasures}
Luz Rello and Miguel Ballesteros. 2015.
\newblock \href {https://doi.org/10.1145/2745555.2746644} {{Detecting readers
  with dyslexia using machine learning with eye tracking measures}}.
\newblock In \emph{Proceedings of the 12th International Web for All
  Conference}, W4A '15, pages 1--8. Association for Computing Machinery.

\bibitem[{Shu et~al.(2003)Shu, Chen, Anderson, Wu, and
  Xuan}]{shu2003properties}
Hua Shu, Xi~Chen, Richard~C. Anderson, Ningning Wu, and Yue Xuan. 2003.
\newblock \href {https://doi.org/10.1111/1467-8624.00519} {{Properties of
  school Chinese: Implications for learning to read}}.
\newblock \emph{Child Development}, 74(1):27--47.

\bibitem[{Vaughn et~al.(2010)Vaughn, Cirino, Wanzek, Wexler, Fletcher, Denton,
  Barth, Romain, and Francis}]{vaughn2010response}
Sharon Vaughn, Paul~T. Cirino, Jeanne Wanzek, Jade Wexler, Jack~M. Fletcher,
  Carolyn~D. Denton, Amy Barth, Melissa Romain, and David~J. Francis. 2010.
\newblock \href {https://doi.org/10.1080/02796015.2010.12087786} {Response to
  intervention for middle school students with reading difficulties: Effects of
  a primary and secondary intervention}.
\newblock \emph{School Psychology Review}, 39(1):3--21.

\bibitem[{Wiechmann et~al.(2022)Wiechmann, Qiao, Kerz, and
  Mattern}]{wiechmann-kerz-2022-measuring}
Daniel Wiechmann, Yu~Qiao, Elma Kerz, and Justus Mattern. 2022.
\newblock \href {https://doi.org/10.18653/v1/2022.acl-long.362} {Measuring the
  impact of (psycho-)linguistic and readability features and their spill over
  effects on the prediction of eye movement patterns}.
\newblock In \emph{Proceedings of the 60th Annual Meeting of the Association
  for Computational Linguistics}, pages 5276--5290. Association for
  Computational Linguistics.

\bibitem[{Wolf et~al.(2020)Wolf, Debut, Sanh, Chaumond, Delangue, Moi, Cistac,
  Rault, Louf, Funtowicz, Davison, Shleifer, von Platen, Ma, Jernite, Plu, Xu,
  Scao, Gugger, Drame, Lhoest, and Rush}]{wolf2020transformers}
Thomas Wolf, Lysandre Debut, Victor Sanh, Julien Chaumond, Clement Delangue,
  Anthony Moi, Pierric Cistac, Tim Rault, Rémi Louf, Morgan Funtowicz, Joe
  Davison, Sam Shleifer, Patrick von Platen, Clara Ma, Yacine Jernite, Julien
  Plu, Canwen Xu, Teven~Le Scao, Sylvain Gugger, Mariama Drame, Quentin Lhoest,
  and Alexander~M. Rush. 2020.
\newblock \href {https://doi.org/10.18653/v1/2020.emnlp-demos.6} {Transformers:
  State-of-the-art natural language processing}.
\newblock In \emph{Proceedings of the 2020 Conference on Empirical Methods in
  Natural Language Processing: System Demonstrations}, pages 38--45.
  Association for Computational Linguistics.

\end{thebibliography}
\bibliographystyle{acl_natbib}

\newpage
\appendix

\section{\citeposs{Pan2014Saccade-targetChinese} dataset}
\label{sec:appendix:data}
Each sentence was composed of seven to 13 words and each word consisted out of one to three characters, with 38 one-character words, 372 two-character words and 22 three-character words. Sentences in which a child blinked while reading a word, except the first and last one, are not included in the final dataset. The set therefore contains the data for between 24 up to 59 sentences for each child.

\section{Reading Measures}
Word-level reading measures used as input for both the baseline models (aggregated over text or subject, respectively) and the neural models. All durations are in ms. Saccade distances refer to distances with respect to x/y-axis coordinates. Landing position refers to character index within a fixated word.
\label{sec:appendix:rm}
\begin{itemize}[noitemsep,leftmargin=*]
\item Horizontal location of fixation on screen
\item Total gaze duration (sum of all fixations landing on the word before moving away from it)
\item Landing position of first fixation within the word
\item Landing position of last fixation within the word 
\item Duration of first fixation 
\item Duration of outgoing saccade
\item Horizontal distance of outgoing saccade
\item Vertical distance of outgoing saccade
\item Total distance of outgoing saccade
\item Duration of incoming saccade
\item Horizontal distance of incoming saccade
\item Vertical distance of incoming saccade
\end{itemize}

\section{Hyperparameter tuning}
\begin{table}[ht!]
  \centering
  \small
  \begin{tabular}{lll}
   \toprule
 \textbf{Model} & \textbf{Hyperparameter} & \textbf{Range}  \\
     \hline\\[-1.5ex]
     \multirow{3}{*}{\textit{Both}} & Batch size & $[8, 16, 32, 64, 128]$   \\
     & Learning rate & $15\times \mathcal{U}\sim(1e^{-5}, 1e^{-1})$  \\
     & Decision boundary & $20\times \mathcal{U}\sim(0.35, 0.65)$  \\
     \midrule
     \multirow{1}{*}{\textit{LSTM}} & Hidden layer size & $[10, 20, \dots, 70]$  \\
     \midrule
     \multirow{8}{*}{\textit{CNN}}
     & C1 \# channels & $[5, 10, \dots, 30]$  \\
     & C1  kernel  & $[3, 5]$ \\
     & C1  pooling  & $[\text{average, max}]$  \\
     & C2 \# channels & $[10, 20, \dots, 50]$  \\
     & C2  kernel  & $[3, 5]$  \\
     & C2  pooling  & $[\text{average, max}]$  \\
     & L1 size & $[10, 20, \dots, 60]$ \\
     & dropout & $[0.1, 0.2, \dots 0.7]$ \\
    \bottomrule
  \end{tabular} 
  \caption{Hyperparameter space for LSTMs and CNNs. \looseness=-1}
  \label{tab:hyperparameter-space}
\end{table}

\end{document}